\documentclass[letterpaper]{article}
\usepackage{aaai20}
\usepackage[hyphens]{url}
\usepackage{times, graphicx}
\usepackage{balance}

\makeatletter
    \let\@internalcite\cite
    \def\cite{\def\citeauthoryear##1##2{##1, ##2}\@internalcite}
    \def\shortcite{\def\citeauthoryear##1{##2}\@internalcite}
    \def\@biblabel#1{\def\citeauthoryear##1##2{##1, ##2}[#1]\hfill}
\makeatother

\title{Abstractive and mixed summarization for long-single documents}

\author{
Roger Barrull Soler \\ 
University of Colorado, Colorado Springs \\ 
rbarrull@uccs.edu 
\and
Jugal Kalita \\ 
University of Colorado, Colorado Springs \\ 
jkalita@uccs.edu
}

\begin{document}
\author{Roger Barrull Soler\\
University of Colorado, Colorado Springs \\
Autonomous University of Barcelona \\
College of Engineering and Applied Sciences \\
rbarrull@uccs.edu 
\And
Jugal Kalita \\ 
University of Colorado, Colorado Springs \\
College of Engineering and Applied Sciences \\
jkalita@uccs.edu
}
	\maketitle
	\begin{abstract}
		The lack of diversity in the datasets available for automatic summarization of documents has meant that the vast majority of neural models for automatic summarization have been trained with news articles. These datasets are relatively small, with an average size of about 600 words, and the models trained with such data sets see their performance limited to short documents. In order to surmount this problem, this paper uses scientific papers as the dataset on which different models are trained. These models have been chosen based on their performance on the CNN/Daily Mail data set, so that the highest ranked model of each architectural variant is selected. 
		In this work, six different models are compared, two with an RNN architecture, one with a CNN architecture, two with a Transformer architecture and one with a Transformer architecture combined with reinforcement learning.
	    The results from this work show that those models that use a hierarchical encoder to model the structure of the document has a better performance than the rest.

	\end{abstract}

	\section{Introduction}
		Summarization is the process of reducing the size of a document while preserving the meaning, and is one of the most researched areas in Natural Language Processing (NLP). Broadly, there are two approaches to summarization techniques on the basis of whether the exact sentences are considered as they appear in the original text or new text is generated. These techniques are called extractive and abstractive, respectively. Extractive summarization has been an extensively researched topic and has reached its maturity stage. So, the focus has shifted towards the more complex abstractive summarization.
		The current state-of-the-art models for abstractive summarization \cite{State} have the limitation that they can only successfully summarize documents only up to a certain length. As a result, when summarizing long documents, we have to truncate the document to the threshold that we trained on. For CNN/Daily Mail, which is the biggest and most used dataset for summarization, this is typically around 600 words.

        An alternative would be to summarize longer text in chunks. However, this limits the coherence of the final summary as semantic information may not flow well between chunks. In addition, finding the right chunking break points is non-trivial, as we have to ensure that at least locally semantic coherent phrases are within the same chunk.
         
        This paper addresses this problem by following the work of Cohen \cite{scientific_papers2} and Nazli\cite{scientific_papers3}. In these papers, they introduced the viability of using scientific papers as a dataset for summarization, considering the abstract as the reference summary for the training.
        
        In order to see which architecture performs better in summarizing long structured documents like scientific papers, the models used in this work correspond to the best models of each architecture on the CNN/Daily Mail dataset. 
        The models are the following.
        \begin{itemize}
        
            \item \textbf{For RNN architecture}: Pointer-Generator-Network with Coverage  \cite{pointer},  and a modification \cite{pointer_disc_aware} that includes a hierarchical encoder and a discourse-aware decoder.
            
            \item \textbf{For CNN architecture}: Convolutional Seq2seq Model \cite{cnn_model}, a hierarchical CNN with copying mechanism.
            \item \textbf{For Transformer architecture}: ProphetNet \cite{prophetnet}, which introduces a novel self-supervised objective named future n-gram prediction and a proposed n-stream self-attention mechanism. And, Transformer Language Model with Extraction \cite{transformer_extractive_step}, which performs a simple extractive step before generating the summary and has a hierarchical encoder architecture.
            \item \textbf{For Transformer + RL architecture}: The model called Sentence Rewriting for Abstractive Summarization \cite{transformer_RL}, which adopts the strategy of extracting salient sentences from a document first and then paraphrasing the selected ones to generate a summary.
        \end{itemize}

        All the models discussed in this work, have been trained with the scientific paper dataset from arXiv. The explanation of each model in this work is conceptual, and if the reader wants to get into the mathematical details of each model, we encourage reading the papers on the original models cited above.

        \section{Top perfoming models for CNN/Daily News}
        
        In Table 1, we list the ranking of the best models in terms of ROUGE for the CNN/Daily Mail dataset and their respective architectures. The ranking is based on comparision found on the website nlpprogress.com. As it can be seen, the ranking is clearly dominated by the models that use the Transformer approach. The models in bold are some of the models that will be compared in this work using the arXiv data set. These models represent the best model for each architecture.
        \vspace{0.5 cm}
        
        \begin{table}[ht]
            \centering
        \begin{tabular}{ |p{1cm}||p{3cm}||p{3cm}|}

        \hline
             Ranking& Model& Architecture\\
          \hline
         \textbf{1}&\textbf{PropheNet}& \textbf{Transformers}\\
         2&PEGASUS\cite{PEGASUS}&Transformers\\
         3&BART\cite{BART}& Transformers \\
         4&T5\cite{T5}& Transformers\\
         5&UniLM\cite{UniLM}&Transformers\\
         \textbf{6}&\textbf{CNN seq2seq}& \textbf{CNN}\\
         7&BertSumExtAbs\cite{BertSumExtAbs}& Transformers\\
         \textbf{8}&\textbf{Bert-ext+abs+RL}&\textbf{Transformers + RL}\\
         ...&...&...\\
         \textbf{22}&\textbf{PGN+Coverage}&\textbf{RNN}\\
         
         \hline
         
    \end{tabular}
    \caption{Ranking extracted from www.nlpprogress.com.}
    \label{tab:caption}
    \end{table}

		\section{Recurrent Neural Networks (RNN)}
			In the past few years, the RNN, which is a type of neural network that can perform computations on sequential data (like sequences of words), has become the standard approach for many NLP tasks. In particular, the encoder-decoder model with attention \cite{attention}, has become popular for summarization.
			However, this model faces two significant problems \cite{problems}.
			\begin{enumerate}
                \item \textbf{The summaries sometimes reproduce factual details}, Producing \textit{He is 5 years old} - instead of - \textit{He is 20 years old}. 
                This is because the model does not copy words from the source text. This happens if there is a word that appears infrequently during training and has a poor word embedding \cite{embedding}, i.e., it is clustered with completely unrelated words. Then the word, from the perspective of the network, is indistinguishable from many other words, and thus impossible to reproduce.
                
                Even if the word has a nice word embedding, the network may still have problems reproducing the word. For example, RNN summarization systems often replace a name with another name, e.g., \textit{Anna} with \textit{Emily}. This is because the word embeddings for female names tend to cluster together, which may cause confusion when attempting to reconstruct the original word.

                 \item \textbf{The summaries sometimes repeat themselves}, e.g., \textit{Germany beat Germany beat Germany beat…} This is because the decoder does not store long-term information in the decoder state. The only information that it receives is the previous summary word. This is why a single repeated word commonly triggers an endless repetitive cycle.
                 \end{enumerate}
                 
        \subsection{Pointer-Generator with Coverage}
            The Pointer-Generator network is a hybrid network that can choose to copy words from the source text using an attention mechanism, while retaining the ability to generate words from a fixed vocabulary. By being able to copy from the source text, it solves the first problem presented earlier. To  tackle  the  second  problem  of  word  repetition, it introduces a technique called Coverage. The idea is to keep track of what has been covered from the source text so far, and penalize the network for attending to the same parts again.
             
             \begin{figure*}[ht]
              \begin{center}
            \includegraphics[width=15cm, height=8cm]{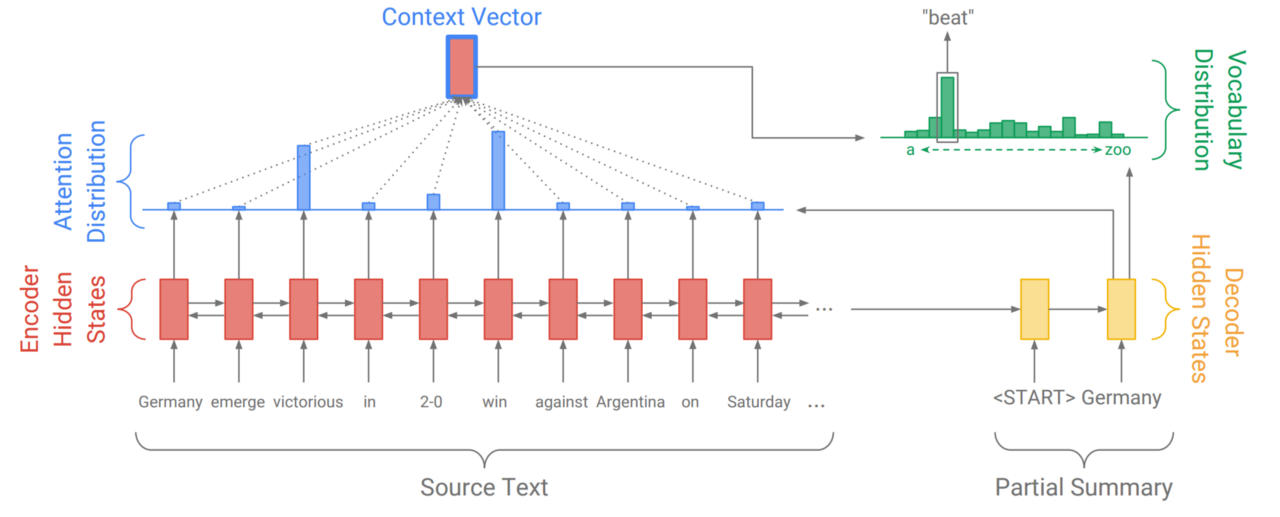}
            \end{center}
            \caption{Architecture of the Pointer Generator Network extracted from the original paper of the model \cite{pointer}. The encoder -in red- is a bidirectional RNN, that reads the source text word-by-word producing a sequence of encoder hidden states that it will pass to the decoder.
            The decoder -in yellow-, which is also an RNN, begins to output a sequence of words that should form a summary. On each step, the decoder receives as input the previous word of the summary, and uses it to update the decoder hidden state.
            The decoder hidden states are used to calculate the attention distribution, which is a probability distribution over the words in the source text. Intuitively, the attention distribution tells the network where to look to help it produce the next word. At the word \textit{Germany}, the decoder is looking at \textit{victorious} and \textit{win} in order to generate the next word.
            Next, the attention distribution is used to produce a weighted sum of the encoder hidden states, known as the context vector. The context vector can be regarded as “what has been read from the source text” on this step of the decoder.
            Finally, the context vector and the decoder hidden states are used to calculate the vocabulary distribution, which is a probability distribution over all the words in a large fixed vocabulary, typically hundreds of thousands of words. The word with the largest probability -on this step, \textit{beat}- is chosen as output.
            Lastly, it calculates the generation probability, which is a scalar between 0 and 1. This represents the probability of generating a word from the vocabulary, versus copying a word from the source.}
            \end{figure*}
             The PGN model with Coverage approach was the state of the art in text summarization from 2017 until models based on Transformers outperformed it in terms of ROUGE \cite{transformers}, the standard metric in the performance evaluation of summaries. However, many people still argue that the PGN model with Coverage produces more readable summaries and with higher coherence when it comes to human evaluation, and there has been some work trying to combine the two \cite{transformer_pgn}.

        \subsection{Discourse-aware attention model}
            The discourse-aware attention model can be seen as the adaptation of the PGN with coverage model for summarizing long-structured documents, e.g., scientific papers.
            
            This model extends the Pointer Generator Network model with coverage by (i) introducing a hierarchical encoder for modeling long documents, and (ii) a discourse-aware decoder that captures the information flow from all discourse sections of the document.
            
            \subsubsection{Hierarchical encoder}
            
            The  RNN encoder is extended to a hierarchical RNN that captures the document discourse structure. It first encodes each discourse section and then encodes the document. The model uses a single bidirectional LSTM layer (following the LSTM formulation of \cite{LSTM}) for both $RNN_{doc}$ and $RNN_{sec}$. The model combines the forward and backward LSTM states to a single state using a simple feed-forward network.
            
            \subsubsection{Discourse-aware decoder}
            
            When humans summarize a long structured document, depending on the domain and the nature of the document, they incorporate important points from different discourse sections of the document. For example, scientific paper abstracts typically include the description of the problem, discussion of the methods, and finally results and conclusion \cite{structure_paper}. Motivated by this observation, the model proposes a discourse-aware attention method. Intuitively, at each decoding timestep, in addition to the words in the document, it also attends to the relevant discourse section (the ``section attention'' block in Figure 2).
            
            \begin{figure*}[ht]
            \begin{center}
            \includegraphics[width=15cm, height=8cm]{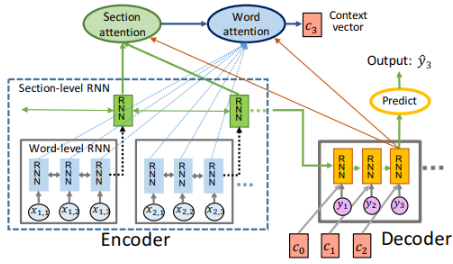}
            \end{center}
            \caption{Overview of the discourse-aware attention model extracted from the original paper of the model \cite{pointer_disc_aware}. In the encoder side at the left, the word-level RNN is shown in blue and the section-level RNN is shown in green. The decoder, seen on the right side, consists of an RNN - in orange- and a “predict” network for generating the summary. At each decoding time step \textit{t} (in the figure \textit{t}=3 is shown), the decoder forms a context vector \textit{ct} which encodes the relevant source context (\textit{c0} is initialized as a zero vector). Then, the section and word attention weights are respectively computed using the green “section attention” and the blue “word attention” blocks. The context vector is used as another input to the decoder RNN and as an input to the “predict” network which outputs the next word using a joint pointer-generator network.}
            \end{figure*}

    \section{Convolutional Neural Networks (CNN)}
    
    RNNs tend to be inefficient as they rely on the previous steps when training. An alternative is the usage of a CNN to create the representations for the source texts. However, traditional CNNs can only encode fixed size contexts. To address this issue, the model, which achieves the highest ROUGE using CNNs, stacks CNN layers over each other. By doing so, the length of the sequence being dealt with can be easily controlled, and each element in the sequence can be computed in parallel. In contrast, an RNN has to preserve the hidden state of the whole past, thus not able to perform parallelization operations.
    
    Multi-layer CNNs create hierarchical representations of the input, i.e., close elements in a sequence interact at lower layers whereas distant elements interact at higher layers. Unlike the chain structure of an RNN, this multi-layer CNN model offers a shortcut to parallelly express long-range sequences

    \subsection{Convolutional Seq2seq Model}
    
    In the evaluated model, the convolutional layer architecture is shared both by the encoder and the decoder, and they calculate intermediate states via the input elements. Each layer consists of a one-dimensional convolution and a non-linearity.

    If a decoder has one layer with kernel width being \textit{k}, its output compresses the information contained in the \textit{k} input elements. To enlarge the length of input elements, the model stacks blocks over one another; for example, stacking 6 blocks with \textit{k}=5 can represent 25 input elements. 
    
    The advantage of the model is that it conducts parallel computation, which is much more efficient than the traditional RNN model, which computes element by element. To represent a sentence with \textit{n} words, CNNs only need \textit{O(n/k)} operations, while RNNs need \textit{O(n)} operations. However, this hierarchical CNN model is not more efficient than traditional CNN models as it has to stack CNN layers to represent a sequence more expressively, while a traditional CNN model only needs one layer to explore a whole sentence.
    
    In the source document, the sentences could be very long; hence, a summary could be drawn from not only the keywords, but also the key sentences. So, firstly, the model applies two CNNs on the source text, at word level and sentence level, respectively. Then, to capture the hierarchical document structure, the model implements a hierarchical attention mechanism applied at both levels simultaneously. It first calculates the word-level attention, which is then re-weighted by its corresponding sentence-level attention.
    In the summarization task, it is common that the named-entities and keywords in the test text are essential to the summary, but in the training data they may actually be rare or even unseen, and therefore these words are out-of-vocabulary (OOV) words. To address the issue, the model introduces a copying mechanism, which enables the seq2seq model to extract the OOV words. It can distinguish the keywords based on their position or syntactic information from the original text, even without knowing much other information.
    
    \subsubsection{Copying Mechanism}
    
    Similar to the Pointer Generator Model, this CNN model introduces a copying mechanism that allows the model to copy words from the source text. This is especially useful when such words are rare words that the model is unable to predict. Each word is assigned a weight in the original document by the copying mechanism. Such a weight is able to evaluate the words importance via a positional attention score, and determine if the word should be copied or not.
    
    \subsubsection{Hierarchical Attention Mechanism}
    
     The model applies the convolutional seq2seq model on both word and sentences levels. Then, it adopts the hierarchical attention mechanism to generate the keywords and the key sentences simultaneously.
    
    \section{Transformers}
    
    The Transformer is a novel architecture that aims to solve sequence-to-sequence tasks by handling long-range dependencies with ease \cite{transformers}. The Transformer relies entirely on self-attention to compute representations of its input and output without using RNNs or CNNs.
    
    Recent work on pretrained language models made significant advances in NLP tasks. BERT \cite{BERT} is a bidirectional encoder that is pretrained by predicting randomly masked tokens in sentences, and by predicting next sentences. GPT \cite{GPT}, GPT-2 \cite{GPT-2} and GPT-3 \cite{GPT-3} are auto-regressive LMs. BART is a pre-trained language model that combines a bidirectional Transformer as an encoder and an auto-regressive Transformer as a decoder. 
    The models based on Transformer used in this work are ProphetNet, a model that is the current state of the art when it comes to abstractive summarization on the CNN/Daily Mail data set, and a model that performs an extractive step before generating the summary.
    
    \subsection{ProphetNet}
    
    ProphetNet is a  pre-trained seq2seq large-scale model that introduces a self-supervised objective named \textbf{future n-gram prediction} and a proposed \textbf{n-stream self-attention mechanism}. Instead of the optimization for one-step ahead prediction in a traditional sequence-to-sequence model, ProphetNet is optimized by \textit{n}-step ahead prediction, which predicts the next \textit{n} tokens simultaneously based on previous context tokens at each time step.
    The ProphetNet model is based on a Transformer encoder-decoder architecture and has two goals:  (a) the model should be able to simultaneously predict the future \textit{n}-grams at each time step in an efficient way during the training phase, and (b) the model should be easily converted to predict the next token only as the original seq2seq model for the inference or finetuning phase. To achieve this, the model extends the two-stream self-attention proposed in XLNet \cite{XLNet} to \textit{n}-stream self-attention. ProphetNet contains a mainstream self-attention mechanism which is the same as the self-attention in the original Transformer. Besides, it introduces \textit{n} extra self-attention predicting streams for future \textit{n}-gram prediction.  During training, the \textit{i}-th predicting stream attends to the hidden states of the main stream to predict the next \textit{i}-th future token, which guarantees every \textit{n} continuous tokens in the target sequence are trained to predict at one time step.
    
     \begin{figure}[ht]
            \begin{center}
            \includegraphics[width=8cm, height=3cm]{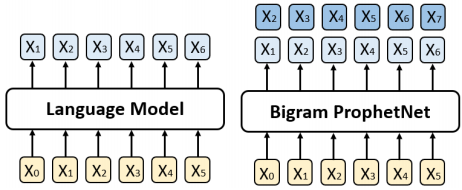}
            \end{center}
            \caption{Traditional language model (left) and ProphetNet (right). The example considers future bigram prediction.}
      \end{figure}
    
   Compared to the original Transformer seq2seq model, ProphetNet introduces four modifications: (a) The novel self-supervised objective called future \textit{n}-gram prediction. (b) The \textit{n}-stream self-attention mechanism, (c) A modified positional embedding, and (d) A mask based auto-encoder denoising task for seq2seq pre-training.

    \begin{figure*}[ht]
            \begin{center}
            \includegraphics[width=14cm, height=10cm]{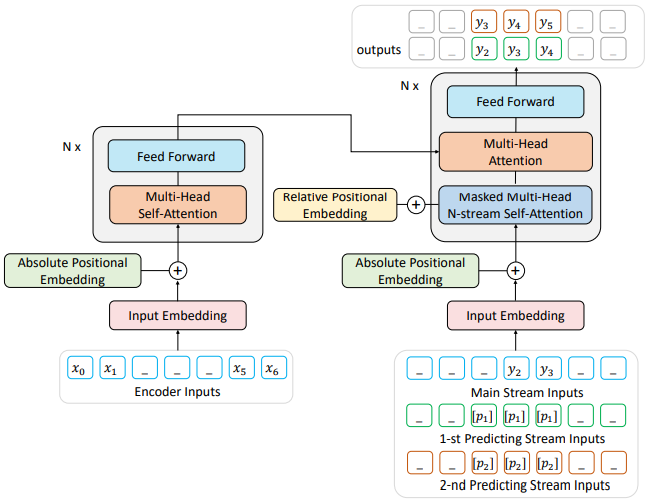}
            \end{center}
            \caption{The architecture of ProphetNet, figure extracted from \cite{prophetnet}. The example considers bigram (n=2). The left part shows the encoder of the ProphetNet, which is the same as the original Transformer encoder. The right part presents the decoder of the ProphetNet, which incorporates the proposed \textit{n}-stream self-attention. For seq2seq pre-training, it can be shown the example of inputs and outputs of the mask based auto-encoder denoising task. The token “-” represents the mask symbol [M]. Note that $x_i$ and $y_i$ are the same in this task for each $i$.}
      \end{figure*}
    \subsection{Transformer Language Model with Extraction}
     
     This model uses a single GPT-like Transformer language model to generate the summary. The interesting thing about this model is that before generating the summary, it does an extractive step, which is then used to condition the Transformer language model on relevant information before being tasked with generating an abstractive summary.
     
     The model comprises two distinct and independently trainable components.
     \begin{itemize}
         \item A hierarchical document representation model that either points to or classifies sentences in a document to build an extractive summary.
         \item A Transformer language model that conditions on the extracted sentences as well as a part of the entire document. 
     \end{itemize}
     
     \begin{figure*}[ht]
            \begin{center}
            \includegraphics[width=15cm, height=8cm]{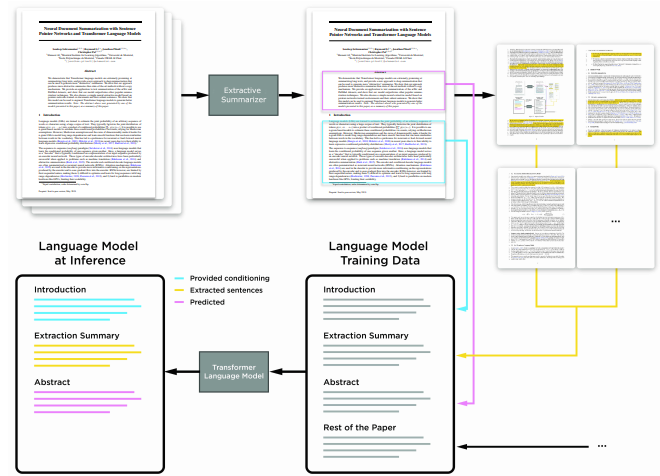}
            \end{center}
            \caption{Proposed model for abstractive summarization of a scientific article by Cohan. Figure extracted from \cite{transformer_extractive_step}. First, a sentence pointer network extracts important sentences from the paper. Next, these sentences are provided along with the whole scientific article to be arranged in the following order: introduction, extracted sentences, abstract and the rest of the paper. A Transformer language model is trained on articles organized in this format. During inference, the introduction and the extracted sentences are given to the language model as context to generate a summary.}
      \end{figure*}
    \subsubsection{Extractive step}  
    
    The model first performs sentence extraction using two different hierarchical document models: one based on pointer networks \cite{pointer_networks}, and the other based on a sentence classifier \cite{sentence_classifier}. This extracts important sentences from the document that can be used to better condition the Transformer language model on relevant information before being tasked with generating a summary

    The model described in the paper, demonstrates that using the ground truth extracted sentences during training and the model extracted sentences at inference performs better than using the model extracted sentences everywhere.

    \subsubsection{Abstractive step: Transformer Language Model (TLM)} 
    
    The model is made up of a single Transformer language model that is trained from scratch. The architecture of this Transformer is identical to the model developed by Radford \cite{GPT-2}.
    The training data is organized for the language model such that the ground-truth summary follows the information used by the model to generate a system summary. This way, it models the joint distribution of document and summary during training, and sample from the conditional distribution of the summary, given the document at inference. 
    When dealing with extremely long documents that may not fit into a single window of tokens seen by a Transformer language model, such as an entire scientific article, the model uses its introduction as a proxy for having enough information to generate an abstract (summary) and uses the remainder of the paper as in domain language model training data (Figure 5).

    \section{Reinforcement Learning (RL)}
   
   Reinforcement Learning (RL) is a widely used learning technique for text summarization. Paulus \cite{RL_paulus} showed that the loss function commonly used in a supervised model is not closely related to the evaluation metric, and introduced an end-to-end RL model that employs the ROUGE metric \cite{rouge} as a rewarder.

    Böhm \cite{better_rewards} highlighted the limitations of ROUGE-based rewarders and proposed neural network-based rewarders to predict the similarity between document and summary. Specifically, the model is trained to predict the similarity score between the document and summaries of various qualities. The pretrained language model BERT is used to encode the input sequences so that the semantics of the two inputs are adequately reflected in the model. 

    \subsection{Summary Level Training of Sentence Rewriting}
    
    The model consists of two neural network modules, an extractor and an abstractor. The extractor encodes a source document and chooses sentences from the document, and then the abstractor paraphrases the selected ones to generate a summary. The model also presents a novel training signal that directly maximizes summary-level ROUGE scores through reinforcement learning.
    
    \subsubsection{Extractor Network}
    
    The extractor is based on the encoder-decoder framework. BERT is adapted for the encoder to exploit contextualized representations from pretrained Transformers. BERT, as the encoder, maps the input sequences to sentence representation vectors.
    
    \subsubsection{Sentence Selection}
    The model uses an LSTM Pointer Network as the decoder to select the extracted sentences based on the sentence representations. The decoder extracts sentences recurrently, producing a distribution over all of the remaining sentence representations, excluding those already selected. Since the sequential model selects one sentence at a time step, the decoder can consider the previously selected sentences. This property is needed to avoid selecting sentences that have overlapping information with the sentences extracted already.
    
     \begin{figure*}[ht]
            \begin{center}
            \includegraphics[width=15cm, height=7cm]{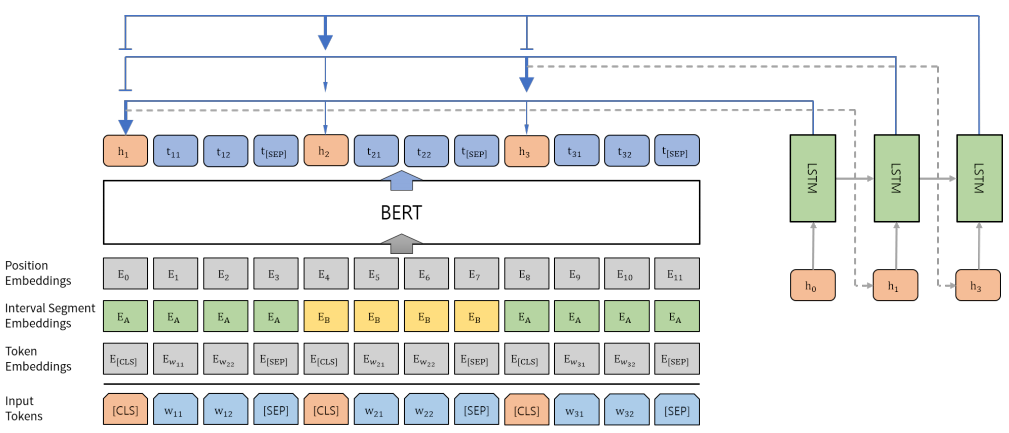}
            \end{center}
            \caption{The overview architecture of the extractor network from the Summary Level Training of Sentence Rewriting model. Figure extracted from \cite{transformer_RL}.}
    \end{figure*}

    \subsubsection{Abstractor Network}
    
    The abstractor network compresses and paraphrases an extracted document sentence to a concise summary sentence. It uses the standard attention based sequence-to-sequence model with the copying mechanism for handling out-of-vocabulary words.
    
    \subsubsection{Training}
        
    To optimize the ROUGE metric directly, the model assumes the extractor as an agent in the reinforcement learning paradigm \cite{RL_agent}. The extractor has a stochastic policy that generates actions (sentence selection) and receives the score of the final evaluation metric (summary-level ROUGE in our case) as the return.

	\section{Datasets}
    The dataset \footnote{Downloadable from https://github.com/armancohan/long-summarization} containing scientific documents for this work was collected from the scientific respository arXiv.org and pre-processed by Cohan and Dernoncourt \cite{pointer_disc_aware}. They removed figures and tables to only preserve the textual information as well as normalized all the math formulas and citation markers with special tokens. The arXiv data set is approximately seven times larger than the popular CNN and Daily Mail dataset of news articles. In Table 2, some statistics of the comparision between the said datasets can be seen. \\
    
    \begin{table}[ht]
            \centering
    \begin{tabular}{ |p{3cm}||p{1cm}||p{1cm}||p{1cm}| }

        \hline
    Dataset& \#docs & avg. doc length (words) & avg. summary length (words)\\
     \hline
     CNN   & 92k    &656&   43\\
     Daily Mail & 219K  & 693   &52\\
     arXiv (this work)&215K & 4938& 220\\

 \hline
\end{tabular}
    \caption{Comparison among the most popular datasets and the one used in this work.}
    \label{tab:caption}
\end{table}

\section{ROUGE}
The set of metrics called ROUGE stand for Recall-Oriented Understudy for Gisting Evaluating \cite{rouge}, and ROUGE is the standard software package used for evaluating automatic summarization. The metrics compare an automatically produced summary against a human-produced one.

\subsection{ROUGE-N}
ROUGE-N is based on the amount of overlap of \textit{n}-grams between the system generated summaries and the reference summaries.
\subsection{ROUGE-L}
It takes into account the longest common subsequences between the system and the reference summaries. It identifies longest co-ocurring subsequence in a sequence of n-grams.

\subsection{Warning}

As stated in www.nlpprogress.com, automatic metrics for summarization such as ROUGE have serious limitations, as outlined below.
\begin{enumerate}
    \item They only assess content selection and do not account for other quality aspects, such as fluency, grammaticality, coherence, etc.

\item To assess content selection, they rely mostly on lexical overlap, although an abstractive summary could express the same content as a reference without any lexical overlap.

\item Given the subjectiveness of summarization and the correspondingly low agreement between annotators, the metrics were designed to be used with multiple reference summaries per input. However, recent datasets such as CNN/DailyMail and Gigaword provide only a single reference.
\end{enumerate}

Therefore, tracking progress and claiming state of the art based only on these metrics is questionable. 

\section{Results}

The results obtained in this work can be seen in Table 3. As it can be observed, the model that uses Transformer language models with an extractive step has the best performance in ROUGE-1 and ROUGE-2, and the Discourse-aware attention model has the best performance in ROUGE-L. The interesting thing about these results is that both models use a hierarchical encoder to capture the document discourse structure. These two models are not even within the top 10 models in the CNN/Daily Mail data set; however, they clearly outperform those top 10 models when the data set is made up of scientific papers.
Regarding the four models selected from their performance on the news datasets, their performance on the arXiv dataset relative to each other is similar to that in the CNN/Daily Mail and there are no big surprises.
This work shows that introducing a hierarchical encoder that models the document is key when it comes to summarize long structured documents such as scientific papers.

\vspace{4mm}
\begin{table}[ht]
\centering
\begin{tabular}{ |p{3cm}||p{1cm}||p{1cm}||p{1cm}||p{1cm}| }

        \hline
    Model& RG-1 & RG-2 & RG-L\\
     \hline
    PGN + coverage  & 32.06    &9.04& 25.16\\
    Disc-aware& 35.8  & 11.05&\textbf{31.80}\\
    CNN seq2seq &35.09 & 12.34& 27.26\\
    ProphetNet &41.01 & 11.19& 23.28\\
    TLM-I+E (G,M)&\textbf{42.43} & \textbf{15.24}& 24.08\\
    Bert + RL &39.00 & 12.09& 22.34\\

    \hline
\end{tabular}
\caption{Comparision between the most popular datasets and the one used in this work.}
\label{tab:caption}
\end{table}

		

	
\balance
	\bibliographystyle{aaai}
	\bibliography{thesis}

\begin{thebibliography}{}

\bibitem[\protect\citeauthoryear{Abigail~See}{2017}]{pointer}
Abigail~See, P.~L.
\newblock 2017.
\newblock Get to the point: Summarization with pointer-generator networks.
\newblock {\em Journal of Machine Learning Research (JMLR)} 2:1--4.

\bibitem[\protect\citeauthoryear{Arman~Cohan}{2017}]{pointer_disc_aware}
Arman~Cohan, F.~D.
\newblock 2017.
\newblock A discourse-aware attention model for abstractive summarization of
  long documents.
\newblock {\em NAACL HLT} 2:2--3.

\bibitem[\protect\citeauthoryear{Bae}{2019}]{transformer_RL}
Bae, S.
\newblock 2019.
\newblock Summary level training of sentence rewriting for abstractive
  summarization.
\newblock {\em EMNLP Workshop on New Frontiers in Summarization} 1:2--5.

\bibitem[\protect\citeauthoryear{Brown}{2020}]{GPT-3}
Brown, T.~B.
\newblock 2020.
\newblock Language models are few-shot learners.

\bibitem[\protect\citeauthoryear{Böhm}{2019}]{better_rewards}
Böhm, F.
\newblock 2019.
\newblock Better rewards yield better summaries: Learning to summarise without
  references.
\newblock {\em arXiv.org}.

\bibitem[\protect\citeauthoryear{Cohan}{2017}]{scientific_papers2}
Cohan, G.
\newblock 2017.
\newblock Contextualizing citations for scientific summarization using word
  embeddings and domain knowledge.
\newblock {\em SIGIR} 1:1--2.

\bibitem[\protect\citeauthoryear{Devlin}{2018}]{BERT}
Devlin, J.
\newblock 2018.
\newblock Bert: Pre-training of deep bidirectional transformers for language
  understanding.
\newblock {\em arXiv.org}.

\bibitem[\protect\citeauthoryear{Dwivedi}{2019}]{problems}
Dwivedi, P.
\newblock 2019.
\newblock Text summarization using deep learning.
\newblock {\em towardsdatascience.com}.

\bibitem[\protect\citeauthoryear{Graves}{2013}]{LSTM}
Graves, A.
\newblock 2013.
\newblock Speech recognition with deep recurrent neural networks.
\newblock {\em In Acoustics, speech and signal processing (icassp)}
  6645–6649.

\bibitem[\protect\citeauthoryear{Jon~Deaton}{2019}]{transformer_pgn}
Jon~Deaton, A.~J.
\newblock 2019.
\newblock Transformers and pointer-generator networks for abstractive
  summarization.
\newblock {\em arXiv.org} 1:3--6.

\bibitem[\protect\citeauthoryear{Lewis}{2019}]{BART}
Lewis, M.
\newblock 2019.
\newblock Bart: Denoising sequence-to-sequence pre-training for natural
  language generation, translation, and comprehension.
\newblock {\em arXiv.org}.

\bibitem[\protect\citeauthoryear{Lin}{2004}]{rouge}
Lin, C.-Y.
\newblock 2004.
\newblock Rouge: A package for automatic evaluation of summaries.
\newblock {\em Association for Computational Linguistics} 1. Text Summarization
  Branches Out:74--81.

\bibitem[\protect\citeauthoryear{Liu}{2019}]{BertSumExtAbs}
Liu, Y.
\newblock 2019.
\newblock Text summarization with pretrained encoders.
\newblock {\em arXiv.org}.

\bibitem[\protect\citeauthoryear{Nallapati}{2016}]{attention}
Nallapati, Z.
\newblock 2016.
\newblock Abstractive text summarization using sequence-to-sequence rnns and
  beyond.
\newblock {\em The SIGNLL Conference on Computational Natural Language Learning
  (CoNLL)} 5:2--4.

\bibitem[\protect\citeauthoryear{Nallapati}{2017}]{sentence_classifier}
Nallapati.
\newblock 2017.
\newblock Summarunner: A recurrent neural network based sequence model for
  extractive summarization of documents.
\newblock {\em arXiv.org}.

\bibitem[\protect\citeauthoryear{Nazli~Goharian}{2017}]{scientific_papers3}
Nazli~Goharian, A.~C.
\newblock 2017.
\newblock Scientific document summarization via citation contextualization and
  scientific discourse.
\newblock {\em International Journal on Digital Libraries (IJDL)} 1:1--3.

\bibitem[\protect\citeauthoryear{Paulus}{2017}]{RL_paulus}
Paulus, R.
\newblock 2017.
\newblock A deep reinforced model for abstractive summarization.
\newblock {\em arXiv.org}.

\bibitem[\protect\citeauthoryear{Radford}{2018}]{GPT}
Radford, A.
\newblock 2018.
\newblock Language models are unsupervised multitask learners.
\newblock {\em arXiv.org}.

\bibitem[\protect\citeauthoryear{Radford}{2019}]{GPT-2}
Radford, A.
\newblock 2019.
\newblock Improving language understanding by generative pre-training.
\newblock {\em https://blog.openai.com/language-unsupervised}.

\bibitem[\protect\citeauthoryear{Raffel}{2019a}]{T5}
Raffel, C.
\newblock 2019a.
\newblock Exploring the limits of transfer learning with a unified text-to-text
  transformer.
\newblock {\em arXiv.org}.

\bibitem[\protect\citeauthoryear{Raffel}{2019b}]{UniLM}
Raffel, C.
\newblock 2019b.
\newblock Exploring the limits of transfer learning with a unified text-to-text
  transformer.
\newblock {\em arXiv.org}.

\bibitem[\protect\citeauthoryear{Som~Gupta}{2018}]{State}
Som~Gupta, S.~G.
\newblock 2018.
\newblock Abstractive summarization: An overview of the state of the art.
\newblock {\em NAACL HLT} 2:1--6.

\bibitem[\protect\citeauthoryear{Suppe}{1998}]{structure_paper}
Suppe, F.
\newblock 1998.
\newblock The structure of a scientific paper.
\newblock {\em Philosophy of Science} 65:381--405.

\bibitem[\protect\citeauthoryear{Sutton}{1998}]{RL_agent}
Sutton, R.~S.
\newblock 1998.
\newblock Introduction to reinforcement learning.
\newblock {\em MIT Press Cambridge.}

\bibitem[\protect\citeauthoryear{Tomas~Mikolov}{2013}]{embedding}
Tomas~Mikolov, K.~C.
\newblock 2013.
\newblock Efficient estimation of word representations in vector space.
\newblock {\em In ICLR Workshop Papers} 3:1--2.

\bibitem[\protect\citeauthoryear{Vaswani}{2017}]{transformers}
Vaswani, A.
\newblock 2017.
\newblock Attention is all you need.
\newblock {\em arXiv.org} 1.

\bibitem[\protect\citeauthoryear{Vinyals}{2015}]{pointer_networks}
Vinyals, O.
\newblock 2015.
\newblock Pointer networks.
\newblock {\em arXiv.org}  2692–2700.

\bibitem[\protect\citeauthoryear{Yan}{2019}]{transformer_extractive_step}
Yan, Y.
\newblock 2019.
\newblock On extractive and abstractive neural document summarization with
  transformer language models.
\newblock {\em arXiv preprint arXiv:1909.03186} 1:1--4.

\bibitem[\protect\citeauthoryear{Yan}{2020}]{prophetnet}
Yan, Y.
\newblock 2020.
\newblock Prophetnet: Predicting future n-gram for sequence-to-sequence
  pre-training.
\newblock {\em MDPI: Applied Sciences} 1:1--5.

\bibitem[\protect\citeauthoryear{Yang}{2019}]{XLNet}
Yang, Z.
\newblock 2019.
\newblock Generalized autoregressive pretraining for language understanding.
\newblock {\em arXiv.org}.

\bibitem[\protect\citeauthoryear{Zhang}{2019a}]{PEGASUS}
Zhang, J.
\newblock 2019a.
\newblock Pegasus: Pre-training with extracted gap-sentences for abstractive
  summarization.
\newblock {\em arXiv.org}.

\bibitem[\protect\citeauthoryear{Zhang}{2019b}]{cnn_model}
Zhang, Y.
\newblock 2019b.
\newblock Abstract text summarization with a convolutional seq2seq model.
\newblock {\em MDPI: Applied Sciences} 1:3--7.

\end{thebibliography}

\end{document}